\documentclass{article}

    \usepackage[final]{neurips_2024}

\usepackage[utf8]{inputenc} % allow utf-8 input
\usepackage[T1]{fontenc}    % use 8-bit T1 fonts
\usepackage{hyperref}       % hyperlinks
\usepackage{url}            % simple URL typesetting
\usepackage{booktabs}       % professional-quality tables
\usepackage{amsfonts}       % blackboard math symbols
\usepackage{nicefrac}       % compact symbols for 1/2, etc.
\usepackage{microtype}      % microtypography
\usepackage{xcolor}         % colors

\usepackage[title]{appendix}

\usepackage{graphicx}
\usepackage{subfigure}
\usepackage{amsmath}
\usepackage{multirow}
\usepackage{multicol}
\usepackage{bbding}
\usepackage{wrapfig}

\usepackage{caption}
\usepackage{amssymb}
\usepackage{footnote}
\usepackage{url}
\usepackage{enumitem}
\usepackage{algorithm}
\usepackage{algorithmic}
\usepackage{listings}
\usepackage{tabularx}
\usepackage {pifont} 
\usepackage{bm}
\usepackage{tablefootnote}
\usepackage{tikz}
\usetikzlibrary{matrix, backgrounds,fit}

\newcommand{\charimage}[1]{\raisebox{-0.25\height}{\includegraphics[height=\baselineskip]{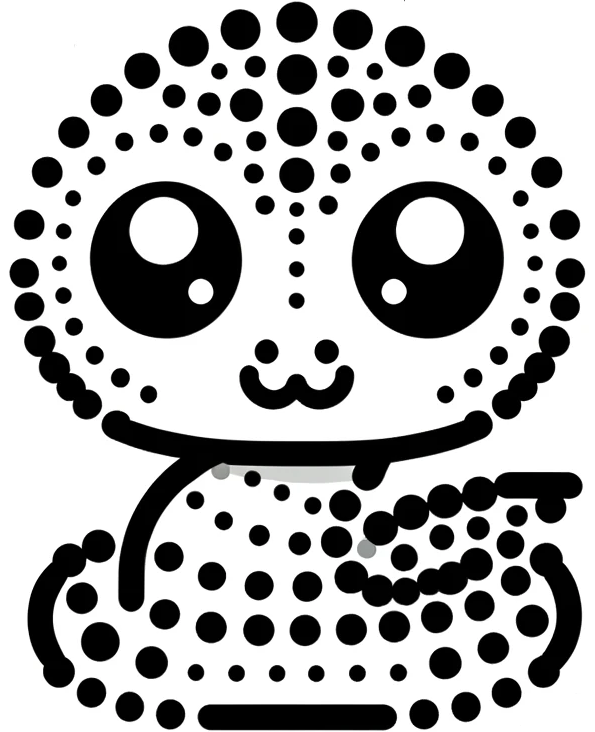}}}

\usepackage{cutwin}
\usepackage{colortbl}
\definecolor{cvprblue}{rgb}{0.21,0.49,0.74}

\definecolor{lightgray}{gray}{0.9}
\definecolor{linecolor}{rgb}{0.82, 0.94, 0.75}
\definecolor{mamba}{RGB}{153, 151, 239}
\definecolor{kaiming-green}{RGB}{57,181,74} % kaiming green
\definecolor{pretty-blue}{RGB}{0, 113, 188}

\def\recon{{\scshape ReCon}}
\def\ours{{PointMamba}}

\definecolor{up}{RGB}{68,169,32}
\definecolor{down}{RGB}{255,0,0}
\definecolor{linecolor}{gray}{.65}

\usepackage{authblk}

\newcommand{\pointmamba}{\textcolor{black}{PointMamba}}

\hypersetup{
    colorlinks=true,
    linkcolor=red,
    filecolor=magenta,      
    urlcolor=magenta,
    citecolor=kaiming-green
}

\title{

\begin{cutout}{0}{0.3cm}{22cm}{1}
\vspace{-30pt}
{\color{white} empty} \protect\linebreak
\protect\linebreak
{\color{white} \qquad } PointMamba: A Simple State Space Model for Point Cloud Analysis
\end{cutout}
}

\author{%
Dingkang Liang$^{1*}$, Xin Zhou$^{1}\thanks{\footnotesize{Equal contribution. $\dagger$ Corresponding author.}}$~, Wei Xu$^{1}$, Xingkui Zhu$^{1}$, 
Zhikang Zou$^{2}$, 

Xiaoqing Ye$^{2}$, Xiao Tan$^{2}$, Xiang Bai$^{1\dagger}$
\\
	$^1$Huazhong University of Science \& Technology, $^2$Baidu Inc.
	\\
	\texttt{\{dkliang, xzhou03, xbai\}@hust.edu.cn}
}

\begin{document}

% \maketitle

\maketitle

{%
\vspace{-9.5mm}
\begin{figure}[H]
\hsize=\textwidth
\centering
\includegraphics[width=0.99\linewidth]{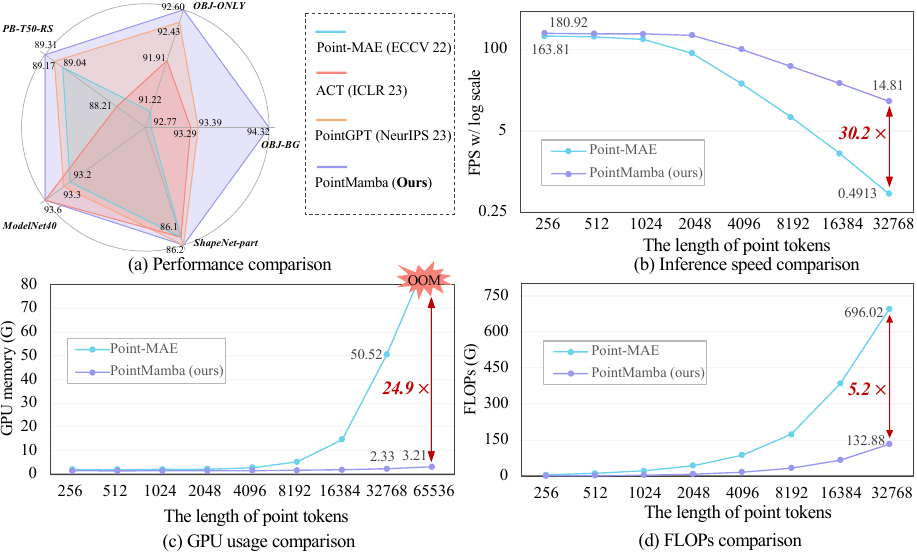}

\caption{
Comprehensive comparisons between our \pointmamba~and its Transformer-based counterparts~\cite{liu2022masked,chen2023pointgpt,dong2022autoencoders}. (a) Without bells and whistles, our \pointmamba~achieve better performance than the representative Transformer-based methods on the various point cloud analysis datasets. (b)-(d) The Transformer presents quadratic complexity, while our \pointmamba~has linear complexity. For example, with the length of point tokens increasing, we significantly reduce GPU memory usage and FLOPs and have a faster inference speed compared to the most convincing Transformer-based method, i.e., PointMAE~\cite{liu2022masked}. }
\vspace{-4mm}
\label{fig:intro}
\end{figure}
}

\begin{abstract}

Transformers have become one of the foundational architectures in point cloud analysis tasks due to their excellent global modeling ability. However, the attention mechanism has quadratic complexity, making the design of a linear complexity method with global modeling appealing. In this paper, we propose \textbf{PointMamba} \charimage{example-image}, transferring the success of Mamba, a recent representative state space model (SSM), from NLP to point cloud analysis tasks. Unlike traditional Transformers, \pointmamba~employs a linear complexity algorithm, presenting global modeling capacity while significantly reducing computational costs. Specifically, our method leverages space-filling curves for effective point tokenization and adopts an extremely simple, non-hierarchical Mamba encoder as the backbone. Comprehensive evaluations demonstrate that \pointmamba~achieves superior performance across multiple datasets while significantly reducing GPU memory usage and FLOPs. This work underscores the potential of SSMs in 3D vision-related tasks and presents a simple yet effective Mamba-based baseline for future research. The code will be made available at \url{https://github.com/LMD0311/PointMamba}.

\end{abstract}

\section{Introduction}

Point cloud analysis is one of the fundamental tasks in computer vision and has a wide range of real-world applications~\cite{shi2020points,wu2023point,chen2023voxelnext}, including robotics, autonomous driving, and augmented reality. It is a challenging task due to the intrinsic irregularity and sparsity of point clouds. To address the issues, there has been rapid progress in deep learning-based methods~\cite{pang2022masked,liu2022masked,qi2017pointnet++,qian2022pointnext}, consistently pushing the performance to the new record.

Recently, Transformers have achieved remarkable progress in point cloud analysis. The key to the Transformer is the attention mechanism, which can effectively capture the relationship of a set of points. By integrating self-supervised learning paradigms with fine-tuning for downstream tasks, these Transformer-based methods have achieved superior performance~\cite{yu2022point, liu2022masked, qi2023recon}. However, the complexity of attention mechanisms is quadratic, bringing significant computational cost, which is not friendly to low-resource devices. Thus, this naturally raises a question: \textit{how to design a simple, elegant method that operates with linear complexity, thereby retaining the benefits of global modeling for point cloud analysis?}

We note the recent advance of the State Space Models (SSMs). As a pioneer, the Structured State Space Sequence Model~\cite{gu2021efficiently} (S4) has emerged as a promising class of architectures for sequence modeling thanks to its strong representation ability and linear-time complexity (achieved by eliminating the need to store the complete context).
Another pioneer, Mamba~\cite{gu2023mamba}, adopts time-varying parameters to the SSM based on S4, proposing an efficient hardware-aware algorithm to enable highly efficient training and inference with dynamic modeling. Recently, a few concurrent methods~\cite{zhu2024vision,liu2024vmamba} successfully transfer the 1D-sequence Mamba from NLP to 2D vision tasks (e.g., image classification and segmentation), achieving similar or surpass the Transformer counterpart~\cite{dosovitskiy2020image} while significantly reducing memory usage. However, regarding the more complex, unstructured data, e.g., 3D point cloud, the effectiveness of Mamba remains unclear. The lack of early exploration of Mamba's potential for point cloud-related tasks hinders further development of its capabilities across the diverse range of applications in this domain.

Inspired by this, this paper aims to unlock the potential of SSM in point cloud analysis tasks, discussing whether it can be a viable alternative to Transformers in this domain. Through a series of pilot experiments, we find that directly using the pioneering SSM, Mamba~\cite{gu2023mamba}, can not achieve ideal performance. We argue that the main inherent limitation comes from the unidirectional modeling employed by the default Mamba, as the context is obtained by compressing the historical hidden state instead of through the interaction between each element.
In contrast, the self-attention of the Transformer is invariant to the permutation of the input elements. 
Given the three-dimensional nature (e.g., unstructured and disordered) of point clouds, using a single scanning process often struggles to concurrently capture dependency information across various directions, which makes it difficult to construct global modeling for the RNN-like modes (e.g., Mamba). 

Therefore, we introduce a simple yet effective Point Cloud State Space Model (denoted as \textbf{PointMamba} \charimage{example-image}) with global modeling and linear complexity. Specifically, to enable Mamba to capture the point cloud structure causally, we first use a point tokenizer to generate two types of point tokens via a point scanning strategy, employing two space-filling curves to scan key points from different directions.  As a result, the unstructured 3D point clouds can be transformed into a regular sequence. The first type of token has local modeling capabilities through sequential encoding, with the latest token holding global sequence information. Consequently, the second type of token can achieve global modeling by containing global information that comes from the first type. Besides, we propose an extremely simple order indicator to maintain the distinct spatial characteristics of different scanning when training, preserving the integrity of the spatial information.  

To make the model as simple as possible, \pointmamba~only employs plain and non-hierarchical Mamba as the backbone to extract features for given serialized point tokens without bells and whistles. We demonstrate that \pointmamba~is very flexible in the pre-training paradigm, where we customize an MAE-like pertaining strategy to provide a good prior, which chooses a random serialization strategy from a pre-defined serialization bank to perform mask modeling. It facilitates the model to exact the general local relationships from different scanning perspectives, better matching the requirement of indirection modeling of mamba. 

Despite no elaborate or complex designs in the model, our \pointmamba~achieves superior performance on various point cloud analysis datasets (Fig.~\ref{fig:intro}(a)). Besides the superior performance, thanks to the linear complexity of Mamba, we show the surprisingly low computational cost\footnote{Note that we removed the tokenizer of both Point-MAE and \pointmamba~, directly fed with a predefined sequence, to better illustrate the structural efficiency.}, as shown in Fig.~\ref{fig:intro}(b)-(c). These notable results underscore the potential of SSM in 3D vision-related tasks.

In conclusion, the contributions of this paper are twofold. \textbf{1)} We introduce the first state space model for point cloud analysis, named \textbf{PointMamba} \charimage{example-image}, which features global modeling with linear complexity. Despite the absence of elaborate or complex structural designs, \pointmamba~demonstrates its potential as an optional model for 3D vision applications.\textbf{ 2)} Our \pointmamba~exhibits impressive capabilities, including structural simplicity (e.g., vanilla Mamba), low computational cost, and knowledge transferability (e.g., support for self-supervised learning).

\section{Related work}                                                   

\subsection{Point Cloud Transformers} 

Vision Transformer (ViT)~\cite{dosovitskiy2020image} has become one of the mainstream architectures in point cloud analysis tasks due to its excellent global modeling ability. Specifically, Point-BERT~\cite{yu2022point} and Point-MAE~\cite{pang2022masked} introduce a standard Transformer architecture for self-supervised learning and is applicable to different downstream tasks. Several works further introduce GPT scheme~\cite{chen2023pointgpt}, multi-scale~\cite{zhang2022point,zha2024towards}, and multi-modal~\cite{dong2022autoencoders,qi2023recon,qi2024shapellm} to guide 3D representation learning.
On the other hand, some researchers~\cite{guo2021pct,zhao2021point,wang2023octformer} focus on modifying the Transformers for point clouds. The PCT~\cite{guo2021pct} conducts global attention directly on the point cloud. Point Transformer~\cite{zhao2021point} applies vector attention~\cite{zhao2020exploring} to perform local attention between each point and its adjacent points. The later Point Transformer series~\cite{wu2022point,wu2023point} further extends the performance and efficiency of the Transformer for different tasks. OctFormer~\cite{wang2023octformer} leverages sorted shuffled keys of octrees to partition point clouds and significantly improve efficiency and effectiveness. 

Standard Transformers can be smoothly integrated into autoencoders using an encoder-decoder design~\cite{he2022masked}, which makes this structure ideal for pre-training and leads to significant performance improvements in downstream point cloud analysis tasks~\cite{yu2022point,pang2022masked,chen2023pointgpt,zha2024lcm,zha2023instance}. However, the attention mechanism has a time complexity of $O(n^2d)$, where $n$ represents the length of the input token sequence and $d$ represents the dimension of the Transformer. This implies that as the input sequence grows, the operational efficiency of the Transformer is significantly constrained.

In this work, we focus on designing a simple State Space Model (SSM) for point cloud analysis without attention while maintaining the global modeling advantages of the Transformer.

\subsection{State Space Models}

Linear state space equations~\cite{gu2020hippo,gu2021combining}, combined with deep learning, offer a compelling approach for modeling sequential data, presenting an alternative to CNNs or Transformers. The Structured State Space Sequence Model~\cite{gu2021efficiently} (S4) leverages a linear state space for contextualization and shows strong performance on various sequence modeling tasks, especially with lengthy sequences. To alleviate computational burden, HTTYH~\cite{gu2022train}, DSS~\cite{gupta2022diagonal}, and S4D~\cite{gu2022parameterization} propose employing a diagonal matrix within S4, maintaining performance without excessive computational costs. The S5~\cite{smith2022simplified} proposes a parallel scan and the MIMO SSM, enabling the state space model to be efficiently utilized and widely implemented. Recently, Mamba~\cite{gu2023mamba} introduced the selective SSM mechanism, a breakthrough achieving linear-time inference and effective training using a hardware-aware algorithm, garnering considerable attention. In the vision domain, Vision Mamba~\cite{liu2024vmamba} compresses the visual representation through bidirectional state space models. VMamba~\cite{zhu2024vision} introduces the Cross-Scan Module, enabling 1D selective scanning in 2D images with global receptive fields. Besides, the great potential of Mamba motivates a series of work across diverse domains, including graph~\cite{wang2024graph,behrouz2024graph}, medical segmentation~\cite{ma2024u,liu2024swin}, video understanding~\cite{li2024videomamba,chen2024video} and generative models~\cite{hu2024zigma,li2024mamba}. 

To the best of our knowledge, there are limited works that study SSMs for point cloud analysis. In this work, we delve into the potential of Mamba in point cloud analysis and propose \pointmamba, which achieves superior performance and significantly reduces computational costs.

\section{Preliminaries}
\label{sec:Preliminaries}
\textbf{State Space Model.} Drawing inspiration from control theory, the State Space Model (SSM) represents a continuous system that maps a state $x_{t}$ to $y_{t}$ through an implicit latent state $h_{t} \in \mathbb{R}^N$. To integrate SSMs into deep models, S4~\cite{gu2021efficiently} defines the system with four parameters ($\boldsymbol{A}, \boldsymbol{B}, \boldsymbol{C},$ and sampling step size $\boldsymbol{\Delta}$). The sequence-to-sequence transformation is defined as:
\begin{align} \label{eq:ssm}
h_{t} = \overline{\boldsymbol{A}} h_{t-1} + \overline{\boldsymbol{B}} x_{t},\qquad y_{t} = \boldsymbol{C}h_{t} +\boldsymbol{D}x_{t},
\end{align}
where $\boldsymbol{C} \in \mathbb{R}^{1 \times N}$ is a project parameter, and $\boldsymbol{D}\in \mathbb{R}^{1 \times N}$ represents a residual connection. The parameters $\overline{\boldsymbol{A}}, \overline{\boldsymbol{B}}$ are defined using the zero-order hold (ZOH) discretization rule:
\begin{align} \label{eq:zoh}
\overline{\boldsymbol{A}} \in \mathbb{R}^{N \times N} = \exp(\boldsymbol{A\Delta}), \qquad
\overline{\boldsymbol{B}} \in \mathbb{R}^{N \times 1} = \left ( \boldsymbol{A\Delta} \right ) ^{-1}\left ( \exp(\boldsymbol{A\Delta}) -\boldsymbol{I}\right )\cdot \boldsymbol{\Delta B}.
\end{align}
 However, the parameter ($\overline{\boldsymbol{A}}, \overline{\boldsymbol{B}}, \boldsymbol{C}, \boldsymbol{\Delta}$) are fixed across all time steps due to the Linear Time-Invariant (LTI) property of SSMs, which limits their capacity to handle varied input sequences.

Recently, Selective SSM (S6) considers parameters $\boldsymbol{B}, \boldsymbol{C}, \boldsymbol{\Delta}$ as functions of the input, effectively transforming the SSM into a time-variant model. Our \pointmamba~adopts a hardware-aware implementation~\cite{gu2023mamba} of S6, showing linear complexity and strong sequence modeling capability.

\textbf{Space-filling curve}. Space-filling curves are paths that traverse every point within a higher-dimensional discrete space while maintaining spatial proximity to a certain degree. Mathematically, they can be defined as a bijective function $\Phi: \mathbb{Z}\to \mathbb{Z}^3$ for point clouds. Our \pointmamba~focuses on the Hilbert space-filling curve~\cite{hilbert1935stetige} and its transposed variant (called Trans-Hilbert), both of which are recognized for effectively preserving locality, ensuring that data points close in $\mathbb{Z}$ space remain close after transformation to $\mathbb{Z}^3$. We note that some methods~\cite{wu2023point,wang2023octformer} utilize space-filling curves to partition the point cloud for capturing spatial contexts, whereas our work mainly focuses on transferring the point clouds to serialization-based sequences and combine with Mamba to implement global modeling. The motivation and objective are different.

\section{PointMamba}

\begin{figure*}[!t]
	\begin{center}
		\includegraphics[width=1.0\linewidth]{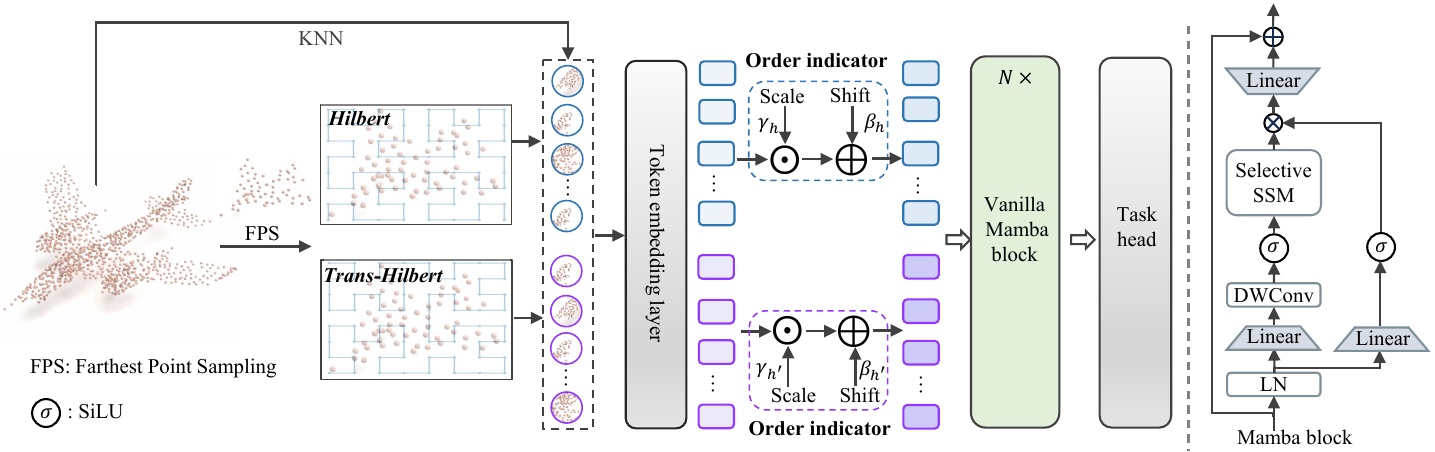}
	\end{center}

	\caption{The pipeline of our \pointmamba. It is simple and elegant, without bells and whistles. We first utilize Farthest Point Sampling (FPS) to select the key points. Then, we propose to utilize two types of space-filling curves, including Hilbert and Trans-Hilbert, to generate the serialized key points. Based on these, the KNN is used to form point patches, which will be fed to the token embedding layer to generate the serialized point tokens. To indicate the tokens generated from which space-filling curve, the order indicator is proposed. The encoder is extremely simple, consisting of $N \times$ plain and non-hierarchical Mamba blocks.}
	\label{fig:pipeline}
\end{figure*}

This paper aims to design a simple yet solid Mamba-based Point cloud analysis method. The pipeline of our method is shown in Fig.~\ref{fig:pipeline}. Starting with an input point cloud, we first sample the key points via Farthest Point Sampling (FPS). Then, a simple space-scanning strategy is applied to reorganize these points, resulting in serialized key points. Under a KNN and lightweight PointNet~\cite{qi2017pointnet}, we obtain the serialized point tokens. Finally, the entire sequence is subsequently processed by a plain, non-hierarchical encoder structure composed of several stacked Mamba blocks. Besides, to provide a good prior for \pointmamba, we propose a serialization-based mask modeling paradigm, which randomly chooses a space-filling curve for serialization and mask, as shown in Fig.~\ref{fig:pretraining}. 

\subsection{The structure of PointMamba}

In this section, we introduce the structure of our \pointmamba. The goal of this paper is to provide a simple yet solid Mamba baseline for point cloud analysis tasks and explore the potential of plain and non-hierarchical Mamba. Thus, in the spirit of \textit{Occam’s razor}, we make the structure as simple as possible without any complex or elaborate design.

\textbf{Point scanning strategy.} Building on the pioneer works~\cite{yu2022point,pang2022masked}, we first utilize the Farthest Point Sampling (FPS) to select the key points. Specifically, given an input point cloud $\boldsymbol{P} \in \mathbb{R} ^{M\times3}$, where $M$ is the number of points, the FPS is applied to sample $n$ key points from the original point cloud $\boldsymbol{P}$, resulting in $\boldsymbol{p} \in \mathbb{R} ^{n\times3}$. In general, the order of the sampling key points $\boldsymbol{p}$ is random, without specific order. This is not a significant problem for the previous Transformer-based methods, as the Transformer is order-invariant when processing sequence data: in the self-attention mechanism, each element at a given position can interact with all other elements in the sequence through attention weights. However, for the selective state space model, i.e., Mamba, we argue that it is hard to model the unstructured point clouds due to the unidirectional modeling. Thus, we propose to leverage the space-filling curves to transform the unstructured point clouds into a regular sequence. Specifically, we choose two representative space-filling curves to scan the key points: the Hilbert curve~\cite{hilbert1935stetige} and its transposed variant, denoted as Trans-Hilbert. Compared with the random sequence, space-filling curves like the Hilbert curve can preserve spatial locality well, i.e., along the scanned 1D serialized point sequence, adjacent key points often have geometrically close positions in 3D space. We argue that this property ensures that the spatial relationships between points are largely maintained, which is crucial for accurate feature representation and analysis in point cloud data. As a complementary, the Trans-Hilbert performs similarly but scans from different clues, which can provide diverse perspectives on spatial locality. By applying Hilbert and Trans-Hilbert to the key points, we obtain two different point serializations, $\boldsymbol{p}_h$ and $\boldsymbol{p}_{h'}$, which will be used to construct point tokens.

\textbf{Point tokenizer }. After obtaining the two serialized key points $\boldsymbol{p}_h$ and $\boldsymbol{p}_{h'}$, we then utilize the KNN algorithm to select $k$ nearest neighbors for each key point, forming $n$ token patches $\boldsymbol{T}_h \in \mathbb{R} ^{n\times k \times 3}$ and $\boldsymbol{T}_{h'} \in \mathbb{R} ^{n\times k \times 3}$ with patch size $k$. To aggregate local information, points within each patch are normalized by subtracting the key point to obtain relative coordinates. We map the unbiased local patches to feature space using a lightweight PointNet~\cite{qi2017pointnet} (point embedding layer), obtaining serialized point tokens $\boldsymbol{E}^h_0\in \mathbb{R} ^{n\times C}$ and $\boldsymbol{E}^{h'}_0 \in \mathbb{R} ^{n\times C}$, where the former is the Hilbert-based and the latter is Trans-Hilbert-based.

\textbf{Order indicator.} Directly fed the two type serialized point tokens $\boldsymbol{E}^h_0\in \mathbb{R} ^{n\times C}$ and $\boldsymbol{E}^{h'}_0 \in \mathbb{R} ^{n\times C}$ into Mamba encoder might cause confusion as $\boldsymbol{E}^h_0$ and $\boldsymbol{E}^{h'}_0$ actually share the same center but with different order. Maintaining the distinct characteristics of these different scanning strategies is important for preserving the integrity of the spatial information. Thus, we propose an extremely simple order indicator to indicate the scanning strategy used. Specifically, the proposed order indicator performs the linear transformation to transfer features into different latent spaces. The formulation can be written as follows:

\begin{equation}
    \boldsymbol{Z}^h_0 = \boldsymbol{E}^h_0 \odot \boldsymbol{\gamma}_h + \boldsymbol{\beta}_h,  \quad  \boldsymbol{Z}^{h'}_0 = \boldsymbol{E}^{h'}_0 \odot \boldsymbol{\gamma}_{h'} + \boldsymbol{\beta}_{h'}, 
    \label{eq:order_indicator}
\end{equation}

where $\boldsymbol{\gamma}_h$/$\boldsymbol{\gamma}_{h'} \in \mathbb{R} ^{C}$ and $\boldsymbol{\beta}_h$/$\boldsymbol{\beta}_{h'} \in \mathbb{R} ^{C}$ refer to the scale and shift factors, respectively. $\odot$ is the dot product. We then concat $\boldsymbol{Z}^h_0$ and $\boldsymbol{Z}^{h'}_0$, resulting in $\boldsymbol{Z}_0 \in \mathbb{R} ^{2n\times C}$.

\textbf{Mamba encoder.} After obtaining the token $\boldsymbol{Z}_0$, we will feed it into the encoder, containing \textit{N} $\times$ Mamba block, to extract the feature. 
Specifically, for each Mamba block, layer normalization ($\mathrm{LN}$), Selective SSM, depth-wise convolution~\cite{chollet2017xception} ($\mathrm{DW}$), and residual connections are employed. A standard Mamba layer is shown in Fig.~\ref{fig:pipeline}, and the output can be summarized as follows:
\begin{equation}
\begin{aligned}
&\boldsymbol{Z}'_{l-1} =  \mathrm{LN} \left ( \boldsymbol{Z}_{l-1} \right ), \qquad\quad\;  \boldsymbol{Z}'_{l}=\sigma \left ( \mathrm{DW}\left ( \mathrm{Linear}\left (  \boldsymbol{Z}'_{l-1}\right )  \right )   \right ) \\
&\boldsymbol{Z}''_{l}=\sigma \left (  \mathrm{Linear}\left (  \boldsymbol{Z}'_{l-1}  \right )   \right ) , ~~\enspace \boldsymbol{Z}_{l}=\mathrm{Linear}\left (\mathrm{SelectiveSSM} \left ( \boldsymbol{Z}'_{l} \right ) \odot \boldsymbol{Z}''_{l}\right ) + \boldsymbol{Z}_{l-1}
\end{aligned} 
\end{equation}

\begin{wrapfigure}{r}{0.49\textwidth}
    % \vspace{-13pt}
    \centering
    \includegraphics[width=0.49\textwidth]{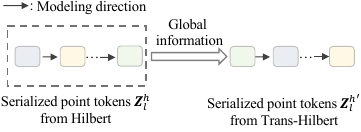}
    \caption{An intuitive illustration of global modeling from \pointmamba. }
    \label{fig:global_modeling}
\end{wrapfigure}
$\boldsymbol{Z}_l \in \mathbb{R} ^{2n\times C}$ is the output of the $l$-th block, and $\sigma$ indicates $\mathrm{SiLU}$ activation~\cite{hendrycks2016gaussian}. The $\mathrm{SelectiveSSM}$ is the key to the Mamba block, with a detailed description in Sec.~\ref{sec:Preliminaries}. To better understand why the proposed \pointmamba~has global modeling capacity, we provide an intuitive visualization. As shown in Fig.~\ref{fig:global_modeling}, after modeling the first group of point tokens (i.e., Hilbert-based), the accumulated global information can improve the serialization process for the next set of tokens (i.e., Trans-Hilbert-based). This mechanism ensures that each serialized point in the Trans-Hilbert sequence is informed by the entire history of the previously processed Hilbert sequence, thereby enabling a more contextually rich and globally aware modeling process. More discussions can be found in Appendix~\ref{sec:appendix-a2}.

In our study, we show that even a very simple Mamba block without specific designs, our \pointmamba~can surpass the various Transformer-based point cloud analysis methods.

\subsection{The serialization-based mask modeling } \label{sec:pretrain}
\begin{figure*}[!t]
	\begin{center}
		\includegraphics[width=0.99\linewidth]{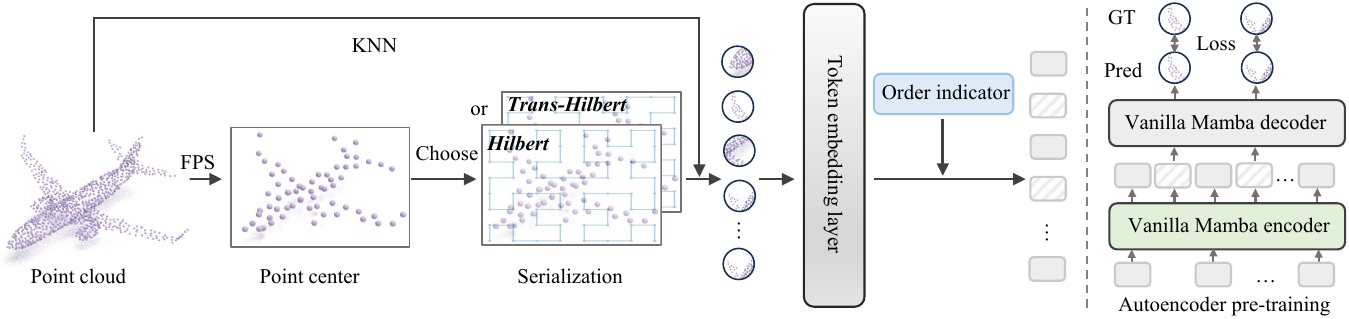}
	\end{center}

	\caption{The details of our proposed serialization-based mask modeling. During the pre-training, we randomly choose one space-filling curve to generate the serialized point tokens for mask modeling, and different serialized point tokens have different order indicators.}
	\label{fig:pretraining}

\end{figure*}

One intriguing characteristic of Transformers-based methods~\cite{liu2022masked,yu2022point,chen2023pointgpt} is their improved performance using the pre-training scheme, especially mask modeling~\cite{he2022masked}. In this paper, considering the unidirectional modeling of Mamba, we customize a simple yet effective serialization-based mask modeling paradigm, as shown in Fig.~\ref{fig:pretraining}.

Specifically, after obtaining the key points, we randomly choose Hilbert or Trans-Hilbert curve to implement serialization in each iteration, resulting in serialization-based key points, i.e., $\boldsymbol{p}_h$ and $\boldsymbol{p}_{h'}$, are obtained from Hilbert and Trans-Hilbert, respectively. Such a scheme allows the model to exact the local relationships from different scanning clues. 
Then, the KNN and the token embedding layer are used to generate the point tokens. To discriminate the point tokens serialized from which space-filling curves, we apply the order indicator to the point tokens, where different serialized point tokens have different order indicators, which are similar to the mentioned Eq.~\ref{eq:order_indicator}. Next, we randomly mask the serialization-based point tokens with a high ratio of 60\%. Then, an asymmetric autoencoder, consisting of several vanilla Mamba blocks, is employed to extract the point feature, and the final layer of the autoencoder utilizes a simple prediction head for reconstruction. To reconstruct masked point patches in coordinate space, we employ a linear head to project the masked token to the shape of the masked input points. The Chamfer Distance~\cite{fan2017point} is then used as the reconstruction loss to recover the coordinates of the points in each masked point patch.

We demonstrate that with such a simple serialization-based mask modeling paradigm, \pointmamba~can easily achieve superior performance.

\section{Experiments}

\subsection{Implementation details}

SSM is new to 3D point cloud analysis, with no existing works detailing the specific implementation. To handle different resolutions of the input point cloud, we divide them into different numbers of patches with a linear scaling (e.g., $M=1024$ input points are divided into $n=64$ point patches), with each patch containing $k=32$ points determined by the KNN algorithm. The \pointmamba~encoder has $N=12$ vanilla Mamba blocks, each Mamba block featuring $C=384$ hidden dimensions. For the pre-training, we utilize ShapeNetCore~\cite{chang2015shapenet} as the dataset, following previous methods~\cite{yu2022point,pang2022masked,chen2023pointgpt}. In addition, we utilize 4 $\times$ Mamba blocks as the decoder to reconstruct the masked point clouds. 

\begin{table}[t]
  \centering
\scriptsize	
\caption{Object classification on the ScanObjectNN dataset~\cite{uy2019revisiting}. We evaluate \ours~on three variants, with PB-T50-RS being the most challenging. Overall accuracy (\%) is reported. Param. denotes the number of tunable parameters during training. {$^\dagger$} indicates that using simple rotational augmentation~\cite{dong2022autoencoders} for training.}
\setlength{\tabcolsep}{1.7mm}
\label{tab:scanobjnn}
\begin{tabular}{lccccccc}
\toprule
Methods & Reference & Backbone & Param. (M)~$\downarrow$& FLOPs (G)~$\downarrow$ & OBJ-BG~$\uparrow$ & OBJ-ONLY~$\uparrow$ & PB-T50-RS~$\uparrow$  \\
\midrule
 \multicolumn{8}{c}{\textit{Supervised Learning Only}}\\
 \midrule
PointNet~\cite{qi2017pointnet} & CVPR 17 & - & 3.5 & 0.5  & 73.3  & 79.2  & 68.0\\
PointNet++~\cite{qi2017pointnet++}& NeurIPS 17 & - & 1.5 & 1.7 & 82.3  & 84.3 & 77.9 \\
PointCNN~\cite{li2018pointcnn}& NeurIPS 18 &- &0.6 & 0.9 & 86.1 & 85.5 & 78.5 \\
DGCNN~\cite{wang2019dynamic}& TOG 19 & - & 1.8 & 2.4 & 82.8  & 86.2  & 78.1 \\
PRANet~\cite{ptpranet}& TIP 21 &- &- & - & - & - & 81.0 \\
MVTN~\cite{hamdi2021mvtn}& ICCV 21 & -  & 11.2 & 43.7 & -     & -     & 82.8 \\
PointNeXt~\cite{qian2022pointnext}& NeurIPS 22  & -  & 1.4 & 1.6 & -     & -  & 87.7 \\
PointMLP~\cite{ma2022rethinking}& ICLR 22  & - &  13.2 & 31.4  & -    & -     & 85.4 \\
RepSurf-U~\cite{ran2022surface}& CVPR 22 & - & 1.5   & 0.8 &  -  & -    & 84.3 \\
ADS~\cite{hong2023attention}& ICCV 23 & - & -  & -  &  - & -   & 87.5 \\
 \midrule
 \multicolumn{8}{c}{\textit{Training from pre-training (Single-Modal)}}\\
 \midrule
  Point-BERT~\cite{yu2022point}& CVPR 22 &Transformer&22.1 & 4.8 & 87.43 & 88.12 & 83.07 \\
  MaskPoint~\cite{liu2022masked}& CVPR 22 & Transformer&22.1 & 4.8 & 89.30 & 88.10 & 84.30 \\
  Point-MAE~\cite{pang2022masked}& ECCV 22 &Transformer& 22.1 & 4.8 & 90.02& 88.29 & 85.18 \\
  Point-M2AE~\cite{zhang2022point}& NeurIPS 22 &Transformer& 15.3 & 3.6 & 91.22 & 88.81 & 86.43 \\
  PointDif~\cite{zheng2023point} & CVPR 24 &Transformer& - & - & 93.29 &91.91 & 87.61 \\
  Point-MAE+IDPT~\cite{zha2023instance}& ICCV 23 &Transformer& 1.7 & 7.2 & 91.22& 90.02 & 84.94 \\
  Point-MAE+DAPT~\cite{zhou2024dynamic}& CVPR 24 &Transformer& 1.1 & 5.0 & 90.88& 90.19 & 85.08 \\
  Point-MAE{$^\dagger$}~\cite{pang2022masked}& ECCV 22 & Transformer& 22.1 & 4.8 & 92.77 & 91.22 & 89.04 \\
  PointGPT-S{$^\dagger$}~\cite{chen2023pointgpt}& NeurIPS 23 & Transformer& 29.2 & 5.7 & 93.39 & 92.43 & 89.17 \\
 \rowcolor{linecolor!40}\ours$^\dagger$~(\textbf{ours})& - &\textbf{Mamba} & \textbf{12.3}& \textbf{3.1} & \textbf{94.32} &\textbf{92.60} & \textbf{89.31}\\

  \midrule
 \multicolumn{8}{c}{\textit{Training from pre-training (Cross-Modal)}}\\
 \midrule
    ACT{$^\dagger$}~\cite{dong2022autoencoders}& ICLR 23 & Transformer& 22.1 & 4.8 & 93.29 & 91.91 & 88.21 \\
    Joint-MAE~\cite{guo2023joint}& IJCAI 23 &Transformer &  - & - & 90.94 & 88.86 & 86.07\\
    I2P-MAE{$^\dagger$}~\cite{zhang2023learning}& CVPR 23 & Transformer &  15.3 & - & 94.15 & 91.57 & 90.11\\
  \recon{$^\dagger$}~\cite{qi2023recon}& ICML 23 & Transformer& 43.6 & 5.3& \textbf{95.18} &\textbf{93.29} & \textbf{90.63} \\
 
 \bottomrule
\end{tabular}
\end{table}
\begin{figure}[t]
    \centering
    \begin{minipage}{0.47\textwidth}
    \renewcommand{\arraystretch}{1.11}
        \centering
        \scriptsize	
        \captionsetup{type=table}
    \caption{Classification on ModelNet40~\cite{wu20153d}. Overall accuracy (\%) is reported. The results are obtained from 1024 points without voting.}
        \label{tab:modelnet}
        \resizebox{\linewidth}{!}{
            \begin{tabular}{lcccc}
            
            \toprule
            Methods & Param. (M)~$\downarrow$& FLOPs (G)~$\downarrow$ & OA (\%)~$\uparrow$  \\
            \midrule
             \multicolumn{4}{c}{\textit{Supervised Learning Only}} \\
             \midrule
            PointNet~\cite{qi2017pointnet} &3.5 &0.5 & 89.2 \\ 
            PointNet++~\cite{qi2017pointnet++} & 1.5 & 1.7  & 90.7 \\ 
            PointCNN~\cite{li2018pointcnn} &0.6 & -  & 92.2 \\ 
            DGCNN~\cite{phan2018dgcnn} &1.8 & 2.4  & 92.9 \\ 
            PointNeXt~\cite{qian2022pointnext} & 1.4 & 1.6  & 92.9 \\
            PCT~\cite{guo2021pct} & 2.9 & 2.3  & 93.2 \\
            OctFormer~\cite{wang2023octformer}& 3.98 & 31.3  & 92.7 \\
            \midrule
             \multicolumn{4}{c}{\textit{with Self-supervised pre-training}} \\
             \midrule
              Point-BERT~\cite{yu2022point}  &22.1 & 2.3  & 92.7 \\
              MaskPoint~\cite{liu2022masked}  &22.1 & 2.3 & 92.6\\
              Point-M2AE~\cite{zhang2022point} & 12.8 & 4.6 & 93.4 \\
              Point-MAE~\cite{pang2022masked}  & 22.1 & 2.4  & 93.2 \\
              PointGPT-S~\cite{chen2023pointgpt} & 29.2 & 2.9 & 93.3 \\
              ACT~\cite{dong2022autoencoders}  & 22.1 & 2.4  & \textbf{93.6}\\
             \rowcolor{linecolor!40}\ours~(\textbf{ours}) & \textbf{12.3}& \textbf{1.5} & \textbf{93.6}\\
             \bottomrule
            \end{tabular}
    }
    \end{minipage}
    \hfill
    \begin{minipage}{0.5\textwidth}
    \renewcommand{\arraystretch}{1.1}
        \centering
        \scriptsize	
        \captionsetup{type=table}
        \caption{Few-shot learning on ModelNet40~\cite{wu20153d}. Overall accuracy (\%)$\pm$the standard deviation (\%) without voting is reported.}
            \footnotesize
          \setlength{\tabcolsep}{0.9mm}
              \label{tab:fewshot}%
            \resizebox{\linewidth}{!}{
                \begin{tabular}{lcccc}
                \toprule
               \multirow{2.3}{*}{Methods} & \multicolumn{2}{c}{5-way} & \multicolumn{2}{c}{10-way} \\
            \cmidrule{2-5}   & 10-shot & 20-shot & 10-shot & 20-shot \\
            
                \midrule
                \multicolumn{5}{c}{\textit{Supervised Learning Only}} \\
    \midrule
    PointNet~\cite{qi2017pointnet} & 52.0$\pm$3.8& 57.8$\pm$4.9 & 46.6$\pm$4.3 & 35.2$\pm$4.8 \\
    PointNet-CrossPoint~\cite{afham2022crosspoint} & 90.9$\pm$1.9& 93.5$\pm$4.4 & 84.6$\pm$4.7 & 90.2$\pm$2.2 \\ 
    DGCNN~\cite{wang2019dynamic} & 31.6$\pm$2.8& 40.8$\pm$4.6 & 19.9$\pm$2.1 & 16.9$\pm$1.5 \\ 
    DGCNN-CrossPoint~\cite{afham2022crosspoint} & 92.5$\pm$3.0& 94.9$\pm$2.1 & 83.6$\pm$5.3 & 87.9$\pm$4.2 \\ 
    \midrule
                \multicolumn{5}{c}{\textit{with Self-supervised pre-training}} \\
                \midrule
                Point-BERT~\cite{yu2022point} & 94.6$\pm$3.1 & 96.3$\pm$2.7 & 91.0$\pm$5.4 & 92.7$\pm$5.1 \\
                MaskPoint~\cite{liu2022masked} & 95.0$\pm$3.7 & 97.2$\pm$1.7 & 91.4$\pm$4.0 & 93.4$\pm$3.5 \\
                Point-MAE~\cite{pang2022masked} & 96.3$\pm$2.5 & 97.8$\pm$1.8 & 92.6$\pm$4.1 & 95.0$\pm$3.0 \\
                Point-M2AE~\cite{zhang2022point} & 96.8$\pm$1.8 & 98.3$\pm$1.4 & 92.3$\pm$4.5 & 95.0$\pm$3.0 \\
                PointGPT-S~\cite{chen2023pointgpt}& 96.8$\pm$2.0 & 98.6$\pm$1.1 & 92.6$\pm$4.6 & 95.2$\pm$3.4 \\
                ACT~\cite{dong2022autoencoders}  & 96.8$\pm$2.3 & 98.0$\pm$1.4 & \textbf{93.3}$\pm$4.0 & \textbf{95.6}$\pm$2.8 \\
                \rowcolor{linecolor!40}\ours~(\textbf{ours}) & \textbf{96.9}$\pm$2.0 & \textbf{99.0}$\pm$1.1 & 93.0$\pm$4.4 & \textbf{95.6}$\pm$3.2 \\
                \bottomrule
                \end{tabular}
             } 
    \end{minipage}
\end{figure}

\subsection{Compared with Transformer-based counterparts}

This paper aims to unlock the potential of Mamba in point cloud tasks, discussing whether it can be a viable alternative to Transformers. Thus, in the following experiments, \textbf{we mainly compare with the state-of-the-art vanilla Transformer-based point cloud analysis methods.}

\textbf{Real-world object classification on ScanObjectNN.} 
ScanObjectNN~\cite{uy2019revisiting} is a challenging 3D dataset comprising about 15,000 objects across 15 categories, scanned from real-world indoor scenes with cluttered complexity backgrounds. As shown in Tab.~\ref{tab:scanobjnn}, we conduct experiments on three versions of ScanObjectNN (i.e., OBJ-BG, OBJ-ONLY, and PB-T50-RS), each with increasing complexity. When compared with the most convincing Transformer-based method, i.e., Point-MAE~\cite{liu2022masked}, \pointmamba~surpasses it by 1.55\%, 1.38\% and 0.27\% on OBJ-BG, OBJ-ONLY, and PB-T50-RS respectively while using less computational costs. Besides, we also outperform the SOTA PointGPT-S~\cite{chen2023pointgpt} by 0.93\%, 0.17\%, 0.14\% across three variants on a comparable scale setting. Note that our method follows \textit{Occam's Razor}, without auxiliary tasks like generation during fine-tuning used in PointGPT~\cite{chen2023pointgpt}. Furthermore, compared to cross-modal learning methods~\cite{dong2022autoencoders,qi2023recon} that use additional training data (cross-modal information) or teacher models, which is not a fair comparison, our \pointmamba~still maintains highly competitive. We mainly want to introduce a new Mamba-based point cloud analysis methods paradigm. Although using some complex designs can bring improvement, they might be heuristics. More importantly, these heuristic designs will decrease the objectivity of the evaluation of our method.

\textbf{Synthetic object classification on ModelNet40.} 
ModelNet40~\cite{wu20153d} is a pristine 3D CAD dataset consisting of 12,311 clean samples across 40 categories. As shown in Tab.~\ref{tab:modelnet}, we report the overall accuracy without adopting the voting strategy. The proposed \pointmamba~achieves the best results compared with various self-supervised Transformer-based methods~\cite{liu2022masked,yu2022point,chen2023pointgpt}. In particular, \pointmamba~surpasses Point-MAE~\cite{liu2022masked} and PointGPT-S~\cite{chen2023pointgpt} by 0.4\% and 0.3\% respectively. It is worth noting that the single-modal-learned \pointmamba~achieves comparable results with the cross-modal-based ACT~\cite{dong2022autoencoders} while significantly reducing parameters and FLOPs about 44\% and 38\%, respectively. Additionally, \pointmamba~demonstrates competitive performance against elaborately designed Transformer models like OctFormer~\cite{wang2023octformer}.

\begin{wrapfigure}{r}{0.5\textwidth}
    \vspace{-10pt}
  \centering
\makeatletter\def\@captype{table}\makeatother
\caption{Part segmentation on the ShapeNetPart~\cite{yi2016scalable}. The mIoU for all classes (Cls.) and for all instances (Inst.) are reported.}
\vspace{-5pt}
    \footnotesize
  \setlength{\tabcolsep}{0.5mm}
      \label{tab:segmentation}
      \resizebox{\linewidth}{!}{
    \begin{tabular}{lcc}
    \midrule
    Methods   &Cls. mIoU (\%)~$\uparrow$ & Inst. mIoU (\%)~$\uparrow$ \\
    \midrule
    \multicolumn{3}{c}{\textit{Supervised Learing Only}} \\
    \midrule
    PointNet~\cite{qi2017pointnet}& 80.39 & 83.7 \\
    PointNet++~\cite{qi2017pointnet++} & 81.85 & 85.1 \\
    DGCNN~\cite{wang2019dynamic}  & 82.33 & 85.2 \\
    APES~\cite{wu2023attention} & 83.67 & 85.8\\
    \midrule
    \multicolumn{3}{c}{\textit{with Self-supervised pre-training}} \\
    \midrule
    Transformer~\cite{yu2022point}   & 83.4 & 85.1 \\
    OcCo~\cite{yu2022point}  & 83.4 & 85.1 \\
    MaskPoint~\cite{liu2022masked}  & 84.6 & 86.0 \\
    Point-BERT~\cite{yu2022point}   & 84.1 & 85.6 \\
    Point-MAE~\cite{pang2022masked} & 84.2 & 86.1 \\ 
    PointGPT-S~\cite{chen2023pointgpt}  & 84.1 & \textbf{86.2} \\ 
    ACT \cite{dong2022autoencoders}  & \textbf{84.7} & 86.1 \\ 
    \rowcolor{linecolor!40}\ours~(\textbf{ours}) & 84.4 & \textbf{86.2}\\
    \bottomrule
    \end{tabular}
    \vspace{-15pt}
    }
    \vspace{-15pt}

\end{wrapfigure}
\textbf{Few-shot learning.} We further conduct few-shot experiments on ModelNet40~\cite{wu20153d} to demonstrate our few-shot transfer ability. Consistent with prior studies~\cite{yu2022point}, we utilize the "$n$-way, $m$-shot" setup, where $n \in \{5,10\}$ denotes the category count and $m \in \{10, 20\}$ represents the samples per category. Following standard procedure, we carry out 10 separate experiments for each setting and reported mean accuracy along with the standard deviation. As indicated in Tab.~\ref{tab:fewshot}, our \pointmamba~shows competitive results with limited data, e.g., +1.0\% mean accuracy compared to the cross-modal method ACT~\cite{dong2022autoencoders} on the 5-way 20-shot split.

\textbf{Part segmentation on ShapeNetPart.} Part segmentation on ShapeNetPart~\cite{yi2016scalable} is a challenging task that aims to predict a more detailed label for each point within a sample. As shown in Tab.~\ref{tab:segmentation}, we report mean IoU (mIoU) for all classes (Cls.) and all instances (Inst.). Our \pointmamba~model demonstrates highly competitive performance compared to the Transformer-based counterparts~\cite{liu2022masked,chen2023pointgpt,yu2022point}. These impressive results further prove the potential of SSM in the point cloud analysis tasks.

\begin{figure*}[!t]
    \centering
    \includegraphics[width=1\linewidth]{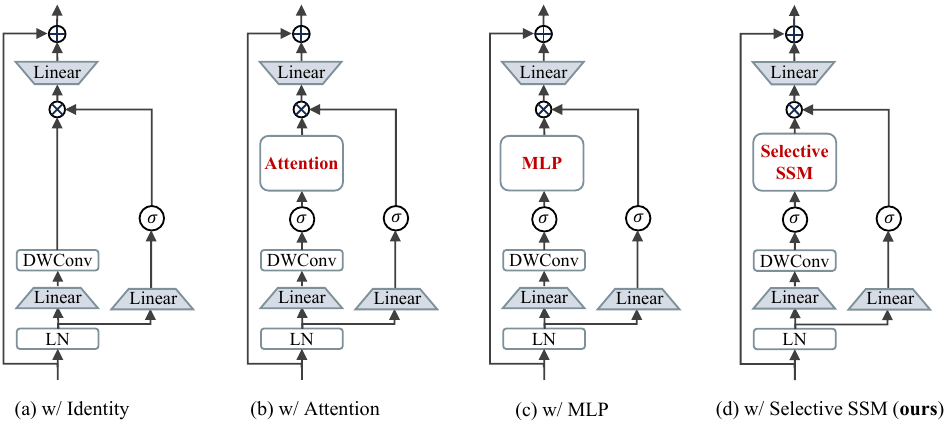}
	\caption{Different variant of \pointmamba. (a) Directly removing the SSM part. (b) Replacing SSM with attention. (c) Replacing SSM with MLP. (d) Ours \pointmamba~with Selective SSM.}
	\label{fig:variant}
\end{figure*}

\begin{figure}[t]
    \centering
    \begin{minipage}{0.46\textwidth}
        \centering
        \small
        \captionsetup{type=table}
        \caption{The effect of each component.}
    \setlength{\tabcolsep}{1.2mm}
    \label{tab:component}
	\centering
	\resizebox{\linewidth}{!}{
	\begin{tabular}{ccccccc}
       \toprule
        Hilbert & Trans-Hilbert& Order indicator & OBJ-BG & OBJ-ONLY\\
        \midrule
        \multicolumn{2}{c}{Random} & -   &92.60 & 90.18\\ 
        \checkmark & - &- &92.94&91.05\\
        -&\checkmark & - &93.46& 91.74\\

        \checkmark & \checkmark & - & 93.80  & 91.91 \\
         \rowcolor{linecolor!40}\checkmark & \checkmark &\checkmark &\textbf{94.32} & \textbf{92.60}\\      

 \bottomrule
        \end{tabular}
	}
    \end{minipage}
    \hfill
    \begin{minipage}{0.48\textwidth}
        \centering
        \small
        \captionsetup{type=table}
        \caption{The effect of different scanning curves.}
    \setlength{\tabcolsep}{2.1mm}
    \label{tab:scanning_curves}
	\centering
	\resizebox{\linewidth}{!}{
	\begin{tabular}{lccc}
         \toprule
        Scanning curve & OBJ-BG & OBJ-ONLY\\
        \midrule
        Random  & 92.60 & 90.18\\
         \rowcolor{linecolor!40}Hilbert and Trans-Hilbert & \textbf{94.32} & \textbf{92.60 }\\
        Z-order and Trans-Z-zorder & 93.29& 90.36 \\
        Hilbert and Z-order & 93.29 & 90.88\\
        Trans-Hilbert and Trans-Z-order & 93.29 & 91.91 \\
        \bottomrule
        \end{tabular}
	}
    \end{minipage}
\end{figure}

\begin{figure}[t]
    \centering
    \begin{minipage}{0.46\textwidth}
        \centering
        \small
        \captionsetup{type=table}
        \caption{The effect of Selective SSM.}
    \setlength{\tabcolsep}{2.4mm}
    \label{tab:ssm}
	\centering
	\resizebox{\linewidth}{!}{
	\begin{tabular}{lccc}
        \toprule
         Setting & Param. & OBJ-BG & OBJ-ONLY \\
        \midrule
         w/ Identity  & 11.4 & 93.80 & 91.57 \\
         w/ Attention & 39.8 & 92.77 & 91.22 \\
        w/ MLP& 18.5 & 93.29 & 91.22 \\

          \rowcolor{linecolor!40} w/ Selective SSM & 12.3 & \textbf{94.32} & \textbf{92.60} \\
        \bottomrule
        \end{tabular}
	}
    \end{minipage}
    \hfill
    \begin{minipage}{0.48\textwidth}
        \centering
        \small
        \captionsetup{type=table}
        \caption{The effect of Order indicator.}

    \setlength{\tabcolsep}{1.5mm}
    \label{tab:indicator}
	\centering
	\resizebox{\linewidth}{!}{
	\begin{tabular}{lccc}
        \toprule
        Setting & OBJ-BG & OBJ-ONLY\\
        \midrule
        None  & 93.80 & 91.91\\ 
        Using the same indicator & 93.29 & 90.19\\
         \rowcolor{linecolor!40} Using different indicator & \textbf{94.32} & \textbf{92.60}\\
        \bottomrule
        \end{tabular}
	}
    \end{minipage}
\end{figure}

\subsection{Analysis and ablation study}

To investigate the architecture design, we conduct ablation studies on ScanObjectNN~\cite{uy2019revisiting} with both pre-training and fine-tuning. Default settings are marked in \colorbox{linecolor!40}{gray}.

\textbf{The structural efficiency.} We first discuss the efficiency of our method. To fully explore the potential of processing the long point tokens (sequence), we gradually increase the sequence length until the GPU (NVIDIA A800 80GB) memory explodes. The comprehensive efficiency comparisons are present in Fig.~\ref{fig:intro}(b)-(d), where Compared with the most convincing Transformer-based method~\cite{liu2022masked}, our \pointmamba~demonstrates significantly improved inference speed and reduce the GPU usage and FLOPs, especially when facing the long sequence. For example, when the length increases to more than 32,768, we outperform PointMAE by 30.2$\times$, 24.9$\times$, and 5.2$\times$ in terms of inference speed, GPU memory, and FLOPs, respectively. More importantly, even presenting impressive efficiency, we still achieve impressive performance on various point cloud analysis datasets.

\textbf{The effect of each component.}
We then study the effectiveness of the proposed components of \pointmamba~as shown in Tab.~\ref{tab:component}. We can make the following observations: 1) Directly utilizing random serialization, \pointmamba~only achieves 92.26\% and 90.18\% overall accuracy on OBJ-BG and OBJ-ONLY, respectively. It is reasonable as Mamba is hard to model the
unstructured point clouds due to its unidirectional modeling. 2) By introducing the locality-preserved Hilbert or Trans-Hilbert scanning, \pointmamba's ability to capture sequence information is enhanced, leading to performance improvements compared to random serialization. Further applying both Hilbert and Trans-Hilbert scanning curves, \pointmamba~surpasses the random serialization by 1.20\% and 1.73\% on two datasets, respectively. 3) By using the order indicator to maintain the distinct characteristics of the two different scanning strategies, we achieve notable improvement, resulting in 94.32\% and 92.60\% on OBJ-BG and OBJ-ONLY, respectively. Note that the order indicator is extremely light (only 1.5k parameters), which will not introduce additional computational costs.

\textbf{The effect of different scanning curves.} 
We further explore the effect of using different scanning curves to construct serialized point tokens. Specifically, we select two widely used space-filling curves, including Hilbert and Z-order, 
along with their transposed variants, i.e., Trans-Hilbert and Trans-Z-order. As listed in Tab.~\ref{tab:scanning_curves}, we empirically find that serializing point clouds with space-filling curves scanning can achieve better performance compared to random sequences. We argue that scanning sequences along a specific pattern of spatial locations offers a more logical sequence modeling order for SSM. We choose the combination of Hilbert and Trans-Hilbert for \pointmamba~due to their superior locality-preserving properties.

\textbf{The effect of Selective SSM.} 
\label{sec:variant}
The key of S6 models or Mamba~\cite{gu2023mamba} is the SSM with the selective mechanism. We prove that as a unidirectional modeling method, SSM can be analogous to masked self-attention, ensuring each position can only attend to previous positions (the detailed proof can be found in the Appendix~\ref{sec:appendix-a1}). Thus, as shown in Tab.~\ref{tab:ssm}, we analyze the effect of selective SSM by removing it (i.e., identity setting) or replace with masked self-attention or MLP (an illustration is shown in Fig.~\ref{fig:variant}). Compared with the identity setting, the selective SSM brings notable improvement, indicating the effectiveness of introducing global modeling from SSM. Note that while a very recent method, MambaOut~\cite{yu2024mambaout}, thinks the SSM of Mamba might negatively impact image classification tasks, our findings demonstrate that this is not the case for point cloud analysis tasks.
Another interesting thing is that when Selective SSM is replaced with masked self-attention, the performance is even lower than that of the identity setting. We argue the main reason is that masked self-attention is hard to combine with the Gated MLP~\cite{shazeer2020glu} used in default Mamba, leading to optimized difficulty, which might need to be explored in the future.

\textbf{Analysis on order indicator.} 
This part analyzes the effect of the order indicator. \pointmamba~applies Hilbert and Trans-Hilbert to recognize the key points, obtaining two types of serialized point tokens $\boldsymbol{E}_{0}^{h}$ and $\boldsymbol{E}_{0}^{h'}$. The order indicator is used to indicate the scanning strategy. As shown in Tab.~\ref{tab:indicator}, using two different order indicators can improve 1.20\% and 1.03\% compared to no indicator on OBJ-BG and OBJ-ONLY, respectively. However, using the same order indicator for both types of sequences without distinguishing between different scanning strategies does not yield positive results.

\subsection{Limitation}

Although \pointmamba~achieves promising results, there are some limitations: 1) We only focus on the point cloud analysis task in this paper while designing a unified Mamba-based foundation model for various 3D vision tasks (e.g., 3D object classification/detection/segmentation) is a more appealing direction. 2) We only use the point clouds as training data while combining them with 2D images or language knowledge to improve the performance, which is also worthy of exploration. We left these in our future work.

\section{Conclusion }

In this paper, we present an elegant, simple Mamba-based method named \pointmamba~for point cloud analysis. \pointmamba~utilizes a space-filling curve-based point tokenizer and a plain, non-hierarchical Mamba architecture to achieve global modeling with linear complexity. Despite its structural simplicity, \pointmamba~delivers state-of-the-art performance across various datasets, significantly reducing computational costs in terms of GPU memory and FLOPs. \pointmamba~success highlights the potential of SSMs, particularly Mamba, in handling the complexities of point cloud data. As a newcomer to point cloud analysis, \pointmamba~is a promising option for constructing 3D vision foundation models, and we hope it can offer a new perspective for the field.

{\small
\bibliographystyle{plain}
\bibliography{references}
}

%%%%%%%%%%%%%%%%%%%%%%%%%%%%%%%%%%%%%%%%%%%%%%%%%%%%%%%%%%%%

\newpage

\begin{appendices}

\section{Theoretically Analysis}

\subsection{Closer look at Selective SSM} \label{sec:appendix-a1}

As described in Sec.~\ref{sec:Preliminaries}, Selective SSM~\cite{gu2023mamba} considers parameters $\boldsymbol{B}, \boldsymbol{C}, \boldsymbol{\Delta}$ in Eq.~\ref{eq:zoh} as functions of the input. To be specific, given the input sequence $\hat{\boldsymbol{x}}=\left [ x_1,\cdots,x_t,\cdots,x_{L} \right ] \in \mathbb{R}^{L\times C} $, the per-time matrices $\boldsymbol{B}_t, \boldsymbol{C}_t, \boldsymbol{\Delta}_t$ can be computed as follows:
\begin{align}\label{eq:per-time}
\boldsymbol{B}_t=L_B(x_t), \quad \boldsymbol{C}_t=L_C(x_t), \quad \boldsymbol{\Delta}_t=\mathrm{softplus}\left ( L_{\Delta}(x_t) \right ),
\end{align}
where $L_B, L_C, L_{\Delta}$ are linear projection layers, and $\mathrm{softplus}(x) = \log (1+e^x)$. The matrices $\overline{\boldsymbol{A}}_t, \overline{\boldsymbol{B}}_t, \boldsymbol{C}_t, \boldsymbol{\Delta}_t$ can be obtained by taking Eq.~\ref{eq:per-time} into Eq.~\ref{eq:zoh}. 

To simplify, we ignore the residual connection $\boldsymbol{D}$ and expand Eq.~\ref{eq:ssm}, the output $\hat{\boldsymbol{y}}=\left [ y_1,\cdots,y_t,\cdots,y_{L} \right ] \in \mathbb{R}^{L\times C} $ can be computed below:
\begin{align}\label{eq:expand}
y_t = \boldsymbol{C}_th_t, \quad h_t=\sum_{i=1}^{t}\left ( \prod_{j=i+1}^{t} \overline{\boldsymbol{A}}_j \right )\overline{\boldsymbol{B}}_i x_i,
\end{align}
which can be further described in matrix form below:
\begin{align}\label{eq:matrix-form}
\begin{bmatrix}
 h_1\\
 h_2\\
 \vdots \\
 h_t\\
\end{bmatrix}
=
\begin{bmatrix}
 \overline{\boldsymbol{B}}_1 & 0 & \cdots & 0\\
 \overline{\boldsymbol{A}}_2\overline{\boldsymbol{B}}_1& \overline{\boldsymbol{B}}_2 & \cdots & 0\\
 \vdots & \vdots & \ddots & \vdots\\
 {\textstyle \prod_{i=2}^{t}} \overline{\boldsymbol{A}}_i\overline{\boldsymbol{B}}_1 & {\textstyle \prod_{i=3}^{t}} \overline{\boldsymbol{A}}_i\overline{\boldsymbol{B}}_2 & \cdots & \overline{\boldsymbol{B}}_t 
\end{bmatrix}
\begin{bmatrix}
 x_1\\
 x_2\\
 \vdots \\
 x_t\\
\end{bmatrix}.
\end{align}
For an intuitive understanding, Eq.~\ref{eq:matrix-form} resembles the self-attention mechanism with a mask $\boldsymbol{M}$, specifically causal self-attention. In this context, $\boldsymbol{M}$ is a lower triangular matrix with elements set to 1. To further exam this, consider the transfer matrix $\boldsymbol{W}$ between $\hat{y}$ and $\hat{x}$, i.e., $\left ( \hat{\boldsymbol{y}} = \boldsymbol{W}\hat{\boldsymbol{x}} \right)$:
\begin{align}
\boldsymbol{W}_{i,j} &= \boldsymbol{C}_i \left ( \prod_{k=j+1}^{i}\overline{\boldsymbol{A}}_k  \right )  \overline{\boldsymbol{B}}_j\\
&= \boldsymbol{C}_i \left ( \prod_{k=j+1}^{i}\exp\left ( \boldsymbol{\Delta_k A}  \right )
 \right ) \overline{\boldsymbol{B}}_j \\
&= \boldsymbol{C}_i  \exp\left ( \sum_{k=j+1}^{i}\boldsymbol{\Delta_k A}  \right ) \overline{\boldsymbol{B}}_j \\
\label{eq:approx}&\approx \boldsymbol{C}_i  \exp\left ( 
\mathop{\sum}_{
        k=j+1
        \atop
        L_{\Delta}\left ( x_k \right )> 0}^{i}L_{\Delta}\left ( x_k \right )  \boldsymbol{A}  \right ) \overline{\boldsymbol{B}}_j,
\end{align}
where $\boldsymbol{W}_{i,j}$ represents the element in the $i$-th row and $j$-th column, the approximation in Eq.~\ref{eq:approx} is done using $\mathrm{ReLU}$ instead of $\mathrm{softplus}$. Consider the notation below:
\begin{align}
\boldsymbol{Q}_{i}: =\boldsymbol{C}_i, \qquad
\boldsymbol{T}_{i,j}=\exp\left ( 
\mathop{\sum}_{
        k=j+1
        \atop
        L_{\Delta}\left ( x_k \right )> 0}^{i} \left (L_{\Delta}\left ( x_k \right )  \boldsymbol{A}\right ) \right ), \qquad
\boldsymbol{K}_j=\left (\overline{\boldsymbol{B}}_j\right )^T.
\end{align}
Thus, the Eq.~\ref{eq:approx} can be simplified to:
\begin{align}
\boldsymbol{W}_{i,j} &\approx \boldsymbol{Q}_{i} \boldsymbol{T}_{i,j} \boldsymbol{K}_{j}^T.
\end{align}
This shows that the Selective SSM captures the influence of $x_i$ and $x_j$ using $\boldsymbol{Q}_{i}$ and $\boldsymbol{K}_{j}$, respectively, while $\boldsymbol{T}_{i,j}$ molding the token significance from $x_i$ to $x_j$. Note that $i\le j$ because $\boldsymbol{W}$ is a lower triangular matrix, indicating a strong relationship with causal self-attention~\cite{brown2020language,ali2024hidden}.

\subsection{Global modeling of PointMamba} \label{sec:appendix-a2}
In this subsection, we explain the global modeling of our \pointmamba. Let's consider the total input sequence as $\left [ \hat{\boldsymbol{l}}_1;\hat{\boldsymbol{l}}_2\right ] = \left [ x_1,,\cdots,x_{l/2}; x_{l/2+1},\cdots,x_l\right ]$ where the sequence has a length $l$ and $l$ is even. $\hat{\boldsymbol{l}}_1$ comes from Hilbert serialization and the other half, $\hat{\boldsymbol{l}}_2$, comes from Trans-Hilbert. The large matrix in Eq.~\ref{eq:matrix-form} can be represented as a partitioned matrix 
$\begin{bmatrix}
  \boldsymbol{X}&\boldsymbol{0} \\
  \boldsymbol{Y}&\boldsymbol{Z}
\end{bmatrix}$ as below:

\begin{equation}
\begin{tikzpicture}
    \matrix [matrix of math nodes,left delimiter={[},right delimiter={]}] (m)
    {
        \overline{\boldsymbol{B}}_1 & & & & & \\
        \vdots &\ddots & & & \\
        {\textstyle \prod_{i=2}^{\frac{l}{2} }} \overline{\boldsymbol{A}}_i\overline{\boldsymbol{B}}_1 & \cdots & \overline{\boldsymbol{B}}_{\frac{l}{2} } & & & \\
        {\textstyle \prod_{i=2}^{\frac{l}{2} +1}} \overline{\boldsymbol{A}}_i\overline{\boldsymbol{B}}_1 & \cdots  &  \overline{\boldsymbol{A}}_{\frac{l}{2} +1}\overline{\boldsymbol{B}}_{\frac{l}{2} } & \overline{\boldsymbol{B}}_{\frac{l}{2} +1} & & \\
        \vdots  & \ddots & \vdots  & \vdots  & \ddots  & \\
        {\textstyle \prod_{i=2}^{l}} \overline{\boldsymbol{A}}_i\overline{\boldsymbol{B}}_1 & \cdots  & {\textstyle \prod_{i=\frac{l}{2} +1}^{l}} \overline{\boldsymbol{A}}_i\overline{\boldsymbol{B}}_{\frac{l}{2} } & {\textstyle \prod_{i=\frac{l}{2} +2}^{l}} \overline{\boldsymbol{A}}_i\overline{\boldsymbol{B}}_{\frac{l}{2} +1} & \cdots & \overline{\boldsymbol{B}}_{l} \\
    };

    \begin{pgfonlayer}{background}
        \fill[gray!25] (m-3-1.south west) rectangle (m-6-3.south east);
    \end{pgfonlayer}

\end{tikzpicture}
\end{equation}

Note that the block $\boldsymbol{Y}$, highlighted in \colorbox{gray!25}{gray}, is associated with both $\hat{\boldsymbol{l}}_1$ (from $\overline{\boldsymbol{B}}$) and $\hat{\boldsymbol{l}}_2$ (from $\overline{\boldsymbol{A}}$), denoted as $\boldsymbol{Y}(\hat{\boldsymbol{l}}_1,\hat{\boldsymbol{l}}_2)$. The blocks $\boldsymbol{X}$ and $\boldsymbol{Z}$ only relate to half of the sequence, denoted as $\boldsymbol{X}(\hat{\boldsymbol{l}}_1)$ and $\boldsymbol{Z}(\hat{\boldsymbol{l}}_2)$, respectively. Thus, the hidden space output $\left [ \hat{\boldsymbol{h}}_1;\hat{\boldsymbol{h}}_2 \right ]^T =\left [ h_1,\cdots,h_{l/2};h_{l/2+1},\cdots,h_{l} \right ]^T$ can be compressed as below:
\begin{align}\label{eq:block-expand}
\begin{bmatrix}
\hat{\boldsymbol{h}}_1\\
\hat{\boldsymbol{h}}_2
\end{bmatrix}
 = 
\begin{bmatrix}
\boldsymbol{X}(\hat{\boldsymbol{l}}_1)\hat{\boldsymbol{l}}_1\\
\boldsymbol{Y}(\hat{\boldsymbol{l}}_1,\hat{\boldsymbol{l}}_2)\hat{\boldsymbol{l}}_2 + \boldsymbol{Z}(\hat{\boldsymbol{l}}_2)\hat{\boldsymbol{l}}_2
\end{bmatrix}.
\end{align}
As in Fig.~\ref{fig:global_modeling} and Eq.~\ref{eq:block-expand}, the serialized points from Trans-Hilbert can receive global information from Hilbert serialization.

\section{Concurrent Related Works}
\begin{table*}[ht]
    \footnotesize
    \setlength{\tabcolsep}{3.mm}
    \centering

  \caption{
Classification performance comparisons with other state space model methods on three variants of the ScanObjectNN~\cite{uy2019revisiting}. All results are reported without voting.
}
\vspace{-5pt}
    \begin{tabular}{lccccccc}
    
    \toprule
    \multirow{2.3}{*}{Method} &\multirow{2.3}{*}{Param. (M)} &\multirow{2.3}{*}{FLOPs (G)} &\multicolumn{3}{c}{ScanObjectNN}\\
		\cmidrule(r){4-6}
	& & &OBJ\_BG & OBJ\_ONLY &PB\_T50\_RS      \\
    \midrule
    Point Cloud Mamba~\cite{zhang2024point} &34.2 &45.0 &- &- &88.10 \\
    % Point Mamba~\cite{} & - & - & - & - & - & 93.4 \\
    Mamba3D~\cite{han2024mamba3d} & 16.9 & 3.9 & 93.12 &92.08 & 88.20 \\
    PoinTramba~\cite{wang2024pointramba} & 19.5 & - & 92.30 &91.30 & 89.10 \\
    \midrule
   \rowcolor{linecolor!40} PointMamba (\textbf{ours}) & \textbf{12.3} & \textbf{3.1} & \textbf{94.32} & \textbf{92.60} & \textbf{89.31}\\

    \bottomrule
    \end{tabular}%
    \label{tab:compare_mamba}

\end{table*}%

Some works on state space models for point cloud analysis appeared recently. This section discusses the differences between these methods and our \pointmamba.

\textbf{Point Cloud Mamba (PCM)}~\cite{zhang2024point} combines an improved Mamba module, i.e., Vision Mamba~\cite{zhu2024vision}, with PointMLP~\cite{ma2022rethinking} (a strong point cloud analysis method), and incorporates consistent traverse serialization at each stage. To enhance Mamba's capability in managing point sequences with varying orders, PCM introduces point prompts that convey the sequence's arrangement rules. While these techniques improve the performance of state space models, they also introduce additional computational overhead and complexity in design.

\textbf{Mamba3D}~\cite{han2024mamba3d} is another recently proposed method. To obtain better global features, Mamba3D introduces an enhanced Vision Mamba~\cite{zhu2024vision} block, which includes both a token forward SSM and a backward SSM that operates on the feature channel. It proposes a Local Norm Pooling block to extract local geometric features.

\textbf{PointTramba}~\cite{wang2024pointramba} introduces a hybrid approach that integrates Transformers and Mamba. It segments point clouds into groups and utilizes Transformers to capture intra-group dependencies, while Mamba models inter-group relationships using a bi-directional, importance-aware ordering strategy.

As shown in Tab.~\ref{tab:compare_mamba}, our \pointmamba~surpasses these concurrent methods by offering superior performance with reduced computational overhead. Note that our method is extremely simple and without complex design, which utilizes the vanilla Mamba block and abstains from incorporating modular designs from other baselines, thereby maintaining simplicity and efficiency in our approach. We believe such a method can better illustrate the potential of SSM in point cloud analysis tasks.

\section{More experimental resutls}

\subsection{Implement details}\label{append-implement}
\vspace{-10pt}
\begin{table}[h]
\setlength{\tabcolsep}{3.4mm}
\caption{Implementation details for pre-training and downstream tasks.}
\label{tab:hyper_params}
\centering
\footnotesize
\begin{tabular}{lcccc}
 \toprule
\multirow{2.3}{*}{Configuration}  & Pre-training  & \multicolumn{2}{c}{Classification}   & Segmentation \\
\cmidrule(r){3-4}
& ShapeNetCore & ModelNet40 & ScanObjectNN & ShapeNetPart \\
 \midrule
 Optimizer & AdamW & AdamW  & AdamW & AdamW\\
 Learning rate & 1e-3 & 3e-4 & 5e-4 & 2e-4 \\
 Weight decay & 5e-2 & 5e-2 & 5e-2  & 5e-2 \\
 Learning rate scheduler & cosine &cosine & cosine & cosine \\
 Training epochs & 300 & 300 & 300 & 300\\
 Warmup epochs & 10 & 10 & 10 & 10\\
 Batch size & 128 & 32 & 32 & 16\\
 \midrule
Num. of encoder layers $N$ & 12 &12 &12 & 12\\
Num. of decoder layers & 4 & -& -& -\\
Input points $M$ & 1024 & 1024 & 2048 & 2048\\
Num. of patches $n$& 64 & 64& 128 &128\\
Patch size $k$ & 32 &32 & 32 &32\\
 \midrule
 Augmentation & Scale\&Trans & Scale\&Trans & Rotation & - \\
\bottomrule[0.95pt]
\end{tabular}
\end{table}

\textbf{Pre-training Details.} The ShapeNetCore dataset~\cite{chang2015shapenet} is used for pre-training, including $\sim$51K clean 3D sample across 55 categories. The 1,024 input points are divided into 64 point patches, with each patch consisting of 32 points. The pre-training process includes 300 epochs, with a batch size of 128. More detail can be found in Tab.~\ref{tab:hyper_params}.

\begin{figure*}[!t]
    \centering
    \includegraphics[width=1\linewidth]{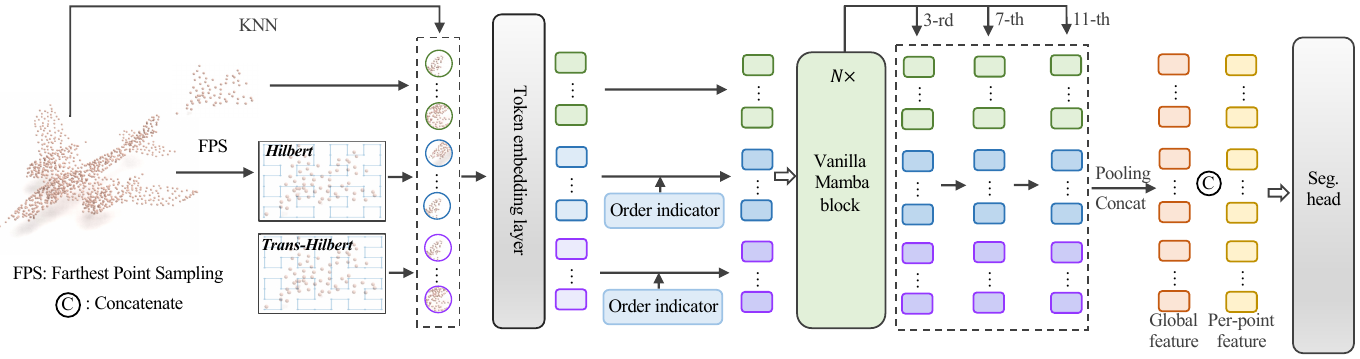}
	\caption{The details of our \pointmamba~for segmentation task.}
	\label{fig:seg-pipelien}
\end{figure*}
\textbf{Downstream tasks Details.} Fig.~\ref{fig:pipeline} shows the pipeline of \pointmamba~for classification tasks. We report the overall accuracy without voting on the challenging ScanObjectNN~\cite{uy2019revisiting} using 2,048 input points, and on ModelNet40~\cite{wu20153d} using 1,024 input points. For segmentation on ShapeNetPart~\cite{yi2016scalable}, as shown in Fig.~\ref{fig:seg-pipelien}, we use random, Hilbert, and Trans-Hilbert serializations, with order indicators applied on Hilbert/Trans-Hilbert serializations. Features from the 3-rd, 7-th, and last layer are pooled as global features after a simple feature fusion. These global features are then concatenated with per-point features and sent to the segmentation head. More detail can be found in Tab.~\ref{tab:hyper_params}.

\subsection{Additional ablation study}\label{append-ablation}
\begin{figure}[t]
    \centering
    \begin{minipage}{0.46\textwidth}
        \centering
        \scriptsize
        \captionsetup{type=table}
        \caption{The effect of masking strategy. The pre-training loss (× 1000) along with fine-tuning accuracy (\%) are reported. }
    \setlength{\tabcolsep}{2.5mm}
    \label{tab:mask}
	\centering
	\vspace{-5pt}
	\resizebox{\linewidth}{!}{
	\begin{tabular}{lccc}
        \toprule
        Masking ratio& Loss & OBJ-BG & OBJ-ONLY\\
        \midrule
        0.4 & 2.01& 92.60&90.70  \\
        \rowcolor{linecolor!40}0.6 & 1.97& \textbf{94.32} & \textbf{92.60}\\
        0.8 & 2.33 & 93.46 & 90.17 \\
        0.9 & 2.00 & 92.43 & 91.05 \\
 \bottomrule
        \end{tabular}
	}
    \end{minipage}
    \hfill
    \begin{minipage}{0.46\textwidth}
        \centering
        \scriptsize
        \captionsetup{type=table}
\caption{The effect of classification token. Fine-tuning accuracy (\%) are reported.}
    \setlength{\tabcolsep}{2.5mm}
    \label{tab:cls}
	\centering
	\vspace{-5pt}
	\resizebox{\linewidth}{!}{
	\begin{tabular}{lccc}
        \toprule
        Methods & OBJ-BG & OBJ-ONLY\\
        \midrule
        Before the sequence & 93.63 & 91.05\\
        After the sequence &  \textbf{94.32} & 90.19\\
        Middle the sequence & 93.98& 90.71 \\
        Max Pool & 93.39& 91.36 \\
       \rowcolor{linecolor!40}Average Pool & \textbf{94.32} & \textbf{92.60}\\
        \bottomrule
        \end{tabular}
	}
    \end{minipage}
\end{figure}

In this section, we do additional ablation studies on several hyper-parameters.

\textbf{Masking strategy for pre-training.} By employing a serialization-based mask modeling paradigm, our \pointmamba~achieves superior performance. To find a proper masking strategy for our method, we compare two types of masking with varying ratios. The block masking~\cite{yu2022point} masks geometrically proximate point cloud patches, leading to a more challenging reconstruction target. In Tab.~\ref{tab:mask}, we experimentally find that masking 60\% of point patches by randomly choosing can achieve good performance.

\textbf{Usage of classification token.} Previous works~\cite{dosovitskiy2020image,yu2022point,pang2022masked} often use a classification token $[\mathtt{CLS}]$ as a global token for classification. As in Tab.~\ref{tab:cls}, we find that without $[\mathtt{CLS}]$ and utilizing only the average pooling of the final block's output yields the best results for \pointmamba.

\section{Qualitative Analysis}

\subsection{Mask modeling visualization}

\begin{figure*}[!t]
    \centering
    \includegraphics[width=1\linewidth]{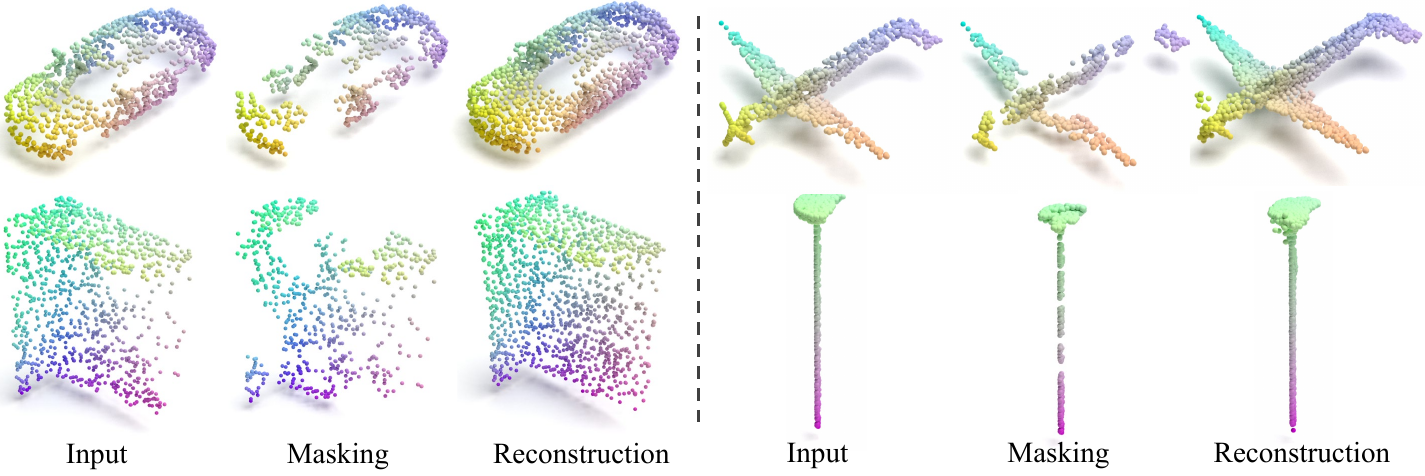}
	\caption{The qualitative results of mask predictions of our \pointmamba~on ShapeNet validation set.}
	\label{fig:pretrain-vis}
\end{figure*}

As in Sec.~\ref{sec:pretrain}, we customize a simple yet effective serialization-based mask modeling paradigm. By randomly masking about 60\% of serialization-based point tokens, an asymmetric vanilla Mamba autoencoder is utilized to extract the point feature, with a simple prediction head for reconstruction. In Fig.~\ref{fig:pretrain-vis}, we present qualitative results of mask modeling on ShapeNet validation set. Despite a 60\% masking ratio, our \pointmamba~effectively reconstructs the masked patches, providing a strong self-supervised knowledge for downstream tasks.

\begin{figure*}[!t]
    \centering
    \includegraphics[width=1\linewidth]{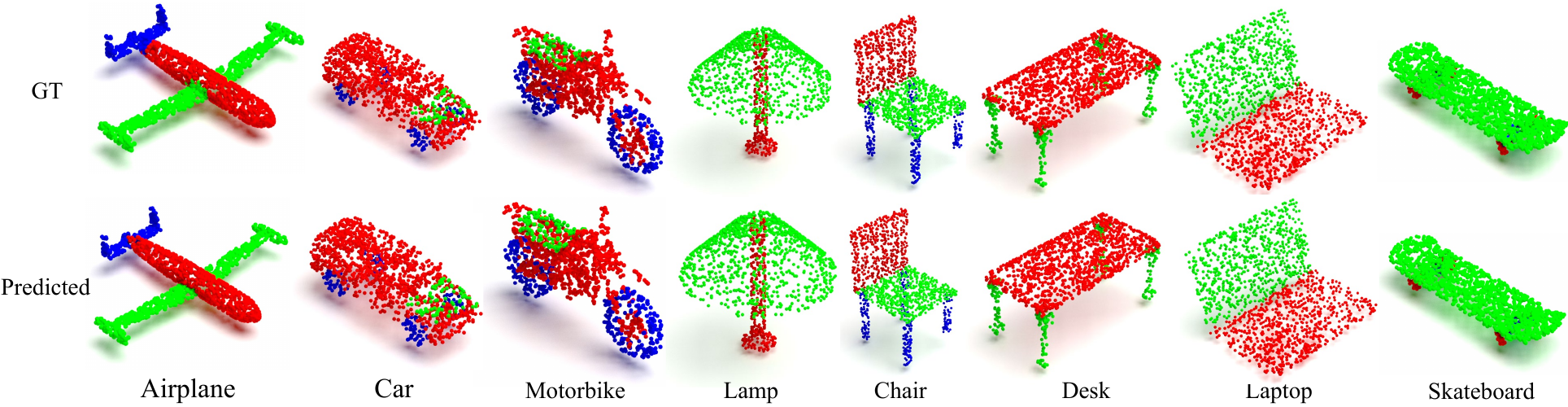}
	\caption{The qualitative results of part segmentation of our \pointmamba~on ShapeNetPart.}
	\label{fig:seg-vis}
\end{figure*}

\subsection{Part segmentation visualization}
In this subsection, we present the qualitative results for part segmentation on the ShapeNetPart validation set, including both the ground truth and the predicted results. As in Fig.~\ref{fig:seg-vis}, our \pointmamba~shows highly competitive results on part segmentation.

\end{appendices}

\end{document}